\documentclass{sig-alternate}
\pdfoutput=1 %

\usepackage{amsmath,amsmath, bm, color,algorithm,array,multirow}
\usepackage[noend]{algorithmic}
\usepackage[
    pdfstartview={FitH},
    bookmarks=true,
    bookmarksnumbered=true,
    bookmarksopen=true,
    bookmarksopenlevel=\maxdimen,
    pdfborder={0 0 0},
    colorlinks=true,
    linkcolor = blue,
    citecolor=blue,
    urlcolor=blue,
    filecolor=blue,
    pdfauthor={Simon Lacoste-Julien, Konstantina Palla, Alex Davies, Gjergji Kasneci, Thore Graepel, Zoubin Ghahramani},
    pdftitle={SiGMa: Simple Greedy Matching for Aligning Large Knowledge Bases},
    pdfdisplaydoctitle=true
]{hyperref}
\usepackage{graphicx}

\toappear{}
\newcommand{\previousversion}[1]{}
\newcommand{\cut}[1]{}
\newcommand{\note}[1]{}
\newcommand{\D}{\displaystyle}
\renewcommand{\tt}[1]{\texttt{#1}}
\newcommand{\ts}[1]{\textsf{#1}}

\newcommand{\KB}{K\!B}

\newcommand{\specialcell}[3][c]{%
  \begin{tabular}[#1]{@{}#2@{}}#3\end{tabular}}

\begin{document}

\title{SiGMa: Simple Greedy Matching\\ for Aligning Large Knowledge Bases}

\numberofauthors{3} %

\author{
\alignauthor
Simon Lacoste-Julien\\
       \affaddr{INRIA - SIERRA project-team}\\
       \affaddr{Ecole Normale Superieure}
       \affaddr{Paris, France}\\
\alignauthor
Konstantina Palla\\
       \affaddr{University of Cambridge}\\
       \affaddr{Cambridge, UK}\\
\alignauthor Alex Davies\\
       \affaddr{University of Cambridge}\\
       \affaddr{Cambridge, UK}\\
\and  %
\alignauthor Gjergji Kasneci\\
       \affaddr{Microsoft Research}\\
       \affaddr{Cambridge, UK}\\
\alignauthor Thore Graepel\\
       \affaddr{Microsoft Research}\\
       \affaddr{Cambridge, UK}\\
\alignauthor Zoubin Ghahramani\\
       \affaddr{University of Cambridge}\\
       \affaddr{Cambridge, UK}\\
}

\maketitle
\begin{abstract}
The Internet has enabled the creation of a growing number of large-scale knowledge bases in a variety of domains containing complementary information. Tools for automatically aligning these knowledge bases would make it possible to unify many sources of structured knowledge and answer complex queries. However, the efficient alignment of \emph{large-scale} knowledge bases still poses a considerable challenge.  Here, we present \textsf{Simple Greedy Matching} (\textsf{SiGMa}), a simple algorithm for aligning knowledge bases with millions of entities and facts. \textsf{SiGMa} is an iterative propagation algorithm which leverages both the structural information from the relationship graph as well as flexible similarity measures between entity properties in a greedy local search, thus making it scalable. Despite its greedy nature, our experiments indicate that \textsf{SiGMa} can efficiently match some of the world's largest knowledge bases with high precision. We provide additional experiments on benchmark datasets which demonstrate that \ts{SiGMa} can outperform state-of-the-art approaches both in accuracy and efficiency.
\end{abstract}

\section{Introduction}
In the last decade, a growing number of large-scale knowledge bases have been created online. Examples of domains include
music, movies, publications and biological data\footnote{Such as \href{http://musicbrainz.org/}{MusicBrainz}, \href{http://www.imdb.com/}{IMDb}, \href{http://www.informatik.uni-trier.de/~ley/db}{DBLP} and \href{http://www.uniprot.org/}{UnitProt}.}. As these knowledge bases sometimes contain both overlapping and complementary information, there has been growing interest in attempting to merge them by \emph{aligning} their common elements. This alignment could have important uses for information retrieval and question answering. For example, one could be interested in finding a scientist with expertise on certain related protein functions -- information which could be obtained by aligning a biological database with a publication one. Unfortunately, this task is challenging to automate as different knowledge bases generally use different terms to represent their entities, and the space of possible matchings grows exponentially with the number of entities.

A significant amount of research has been done in this area -- particularly under the umbrella term of \emph{ontology matching} \cite{choi06survey, kalfoglou03om-state-of-the-art, euzenat07om-book}. An ontology is a formal collection of world knowledge and can take different structured representations. In this paper, we will use the term \emph{knowledge base} to emphasize that we assume very little structure about the ontology (to be specified in Section~\ref{sec:problem}). Despite the large body of literature in this area, most of the work on ontology matching has been demonstrated only on fairly small datasets of the order of a few hundred entities. In particular, Shvaiko and Euzenat~\cite{shvaiko08challenges} identified \emph{large-scale evaluation} as one of the ten challenges for the field of ontology matching.

In this paper, we consider the problem of aligning the \emph{instances} in \emph{large} knowledge bases, of the order of millions of entities and facts, where \emph{aligning} means automatically identifying corresponding entities and interlinking them. Our starting point was the challenging task of aligning the movie database \textsf{IMDb} to the Wikipedia-based \textsf{YAGO}~\cite{suchanek2007WWW}, as another step towards the Semantic Web vision of interlinking different sources of knowledge which is exemplified by the Linking Open Data Initiative\footnote{\href{http://linkeddata.org/}{http://linkeddata.org/}}~\cite{lee08WWW}. Initial attempts to
match \textsf{IMDb} entities to \textsf{YAGO} entities by naively exploiting string and neighborhood information failed, and so we designed \textsf{SiGMa} (\textsf{Simple Greedy Matching}), a scalable greedy iterative algorithm which is able to exploit previous matching decisions as well as the relationship graph information between entities.

The design decisions behind \textsf{SiGMa} were both to be able to take advantage of the combinatorial structure of the matching problem (by contrast with database record linkage approaches which make more independent decisions) as well as to focus on a simple approach which could be scalable. \textsf{SiGMa} works in two stages: it first starts with a small seed matching assumed to be of good quality. Then the algorithm incrementally augments the matching by using \emph{both} structural information and properties of entities such as their string representation to define a modular score function. Some key aspects of the algorithm are that (1) it uses the current matching to obtain structural information, thereby harnessing information from previous decisions; (2) it proposes candidate matches in a \emph{local} manner, from the structural information; and (3) it makes greedy decisions, enabling a scalable implementation. A surprising result is that we obtained accurate large-scale matchings in our experiments despite the greediness of the algorithm.

\paragraph*{Contributions} The contributions of the present work are the following:
\begin{enumerate}
\item We present \textsf{SiGMa}, a knowledge base alignment algorithm which can handle millions of entities. The algorithm is easily extensible with tailored scoring functions to incorporate domain knowledge. It also provides a natural tradeoff between precision and recall, as well as between computation and recall.
  \item In the context of testing the algorithm, we constructed two large-scale partially labeled knowledge base
    alignment datasets with hundreds of thousands of ground truth mappings. We expect these to be a useful resource for the research community to develop and evaluate new knowledge base alignment algorithms.
  \item We provide a detailed experimental comparison illustrating how \textsf{SiGMa} improves over the state-of-the-art. \textsf{SiGMa} is able to align knowledge bases with millions of entities with over 95\% precision in less than two hours (a 50x speed-up over~\cite{suchanek12PARIS}). On standard benchmark datasets, \textsf{SiGMa} obtains solutions with higher F-measure than the best previously published results.
\end{enumerate}

The remainder of the paper is organized as follows. Section~\ref{sec:problem} presents the knowledge base alignment problem with a real-world example as motivation for our assumptions. We describe the algorithm \textsf{SiGMa} in Section~\ref{sec:algorithm}. We evaluate it on benchmark and on real-world datasets in Section~\ref{sec:experiments}, and situate it in the context of related work in Section~\ref{sec:related}.

\section{Aligning Large-Scale Knowledge Bases} \label{sec:problem}

\subsection{Motivating example: \textsf{YAGO} and \textsf{IMDb}}
Consider merging the information in the following two knowledge bases:
\begin{enumerate}
    \item \textsf{YAGO}, a large semantic knowledge base derived from English Wikipedia~\cite{suchanek2007WWW}, WordNet~\cite{wordnet} and GeoNames.\footnote{\href{http://www.geonames.org/}{http://www.geonames.org/}}
 \item \textsf{IMDb}, a large popular online database that stores information about movies.\footnote{\href{http://www.imdb.com/}{http://www.imdb.com/}}
\end{enumerate}
The information in \textsf{YAGO} is available as a long list of triples (called \emph{facts}) that we formalize as:
\begin{equation}
\label{eq:triplet}
\langle e, r, e'\rangle,
\end{equation}
which means that the directed relationship $r$ holds from entity $e$ to entity $e'$, such as $\langle \textrm{John\_Travolta}, \textrm{ActedIn}, \textrm{Grease}\rangle$. The information from \textsf{IMDb} was originally available as several files which we merged into a similar list of triples.  We call these two databases \emph{knowledge bases} to emphasize that we are not assuming a richer representation, such as RDFS \cite{RDFs}, which would distinguish between classes and instances for example. In the language of ontology matching, our setup is the less studied \emph{instance matching} problem, as point\-ed out by Castano et al.~\cite{castano08instanceMatching}, for which the goal is to match concrete instantiations of concepts such as specific actors and specific movies rather than the general actor or movie class.
\previousversion{
}
\textsf{YAGO} comes with an RDFS representation, but not \textsf{IMDb}; therefore we will focus on methods that do not assume or require a class structure or rich hierarchy in order to find a one-to-one matching of instances between \textsf{YAGO} and \textsf{IMDb}.

We note that in the full generality of the ontology matching problem, both the schema and the instances of one ontology are to be related with the ones of the other ontology. Moreover, in addition to the \texttt{isSameAs} (or ``$\equiv$'') relationship that we consider, these matching relationships could be \texttt{isMoreGeneralThan} (``$\supseteq$''), \tt{isLessGeneral\-Than} (``$\subseteq$'') or even \texttt{hasPartialOverlap}. In our example, because the number of relations in the knowledge bases is relatively small (108 in \textsf{YAGO} and 10 in \textsf{IMDb}), we could align the relations manually, discovering six equivalent ones as listed in Table~\ref{tab:relations}. As we will see in our experiments, focussing uniquely on the \texttt{isSameAs} type of relationship between instances of the two knowledge bases is sufficient in the \textsf{YAGO}-\textsf{IMDb} setup to cover most cases. %
The exceptions are rare enough for \textsf{SiGMa} to obtain useful results while making the simplifying assumption that the alignment between the instances is \emph{injective} (1-1).
\begin{table}
\begin{center}
\begin{tabular}{cc}
\hline
\textsf{YAGO} & \textsf{IMDb} \\ \hline
actedIn  & actedIn  \\
directed & directed \\
produced & produced \\
created & composed \\
hasLabel$^*$ & hasLabel$^*$ \\
wasCreatedOnDate$^*$ & hasProductionYear$^*$ \\
\hline
\end{tabular}
\end{center}
\caption{Manually matched relations between \textsf{YAGO} and \textsf{IMDb}. \textnormal{The starred pairs are actually pairs of \emph{properties}, as defined in the text.}\label{tab:relations}}
\end{table}

\paragraph*{Relationships vs. properties} Given our assumption that the alignment is 1-1, it is important to  distinguish between two types of objects which could be present in the list of triples: \emph{entities} vs. \emph{literals}. By our definition, the \emph{entities} will be the only objects that we will try to align -- they will be objects like specific actors or specific movies which have a clear identity. The \emph{literals}, on the other hand, will correspond to a value related to an entity through a special kind of relationship that we will call \emph{property}. The defining characteristic of literals is that it would not make sense to try to align them between the two knowledge bases in a 1-1 fashion. For example, in the \textsf{YAGO} triple $\langle \tt{m1}$, $\tt{wasCreatedOnDate}$, $\tt{1999-12-11}\rangle$, the object \tt{1999-12-11} could be interpreted as a literal representing the value for the property \tt{wasCreated\-On\-Date} for the entity \tt{m1}. The corresponding property in our version of \textsf{IMDb} is \tt{has\-Production\-Year} which has values only at the year granularity (\tt{1999}). The 1-1 restriction would prevent us to align both \tt{1999-12-11} and \tt{1999-12-10} to \tt{1999}. On the other hand, we can use these literals to define a similarity score between entities from the two knowledge bases (for example in this case, whether the year matches, or how close the dates are to each other). We will thus have two types of triples: entity-relationship-entity and entity-property-literal. We assume that the distinction between relationships and properties (which depends on the domain and the user's goals) is easy to make; for example, in the \textsf{Freebase} dataset that we also used in our experiments, the entities would have unique identifiers but not the literals. Figure~\ref{fig:example} provides a concrete example of information presents in the two knowledge bases that we will keep re-using in this paper.

We are now in a position to state more precisely the problem that we address.

\textbf{Definition:} A \emph{knowledge base} $\KB$ is a tuple \\
$(\mathcal{E},\mathcal{L},\mathcal{R},\mathcal{P},\mathcal{F}_R,\mathcal{F}_P)$ where $\mathcal{E}$, $\mathcal{L}$, $\mathcal{R}$ and $\mathcal{P}$ are sets of entities, literals, relationships and properties respectively; $\mathcal{F}_R \subseteq \mathcal{E}\times\mathcal{R}\times\mathcal{E}$ is a set of relationship-facts whereas $\mathcal{F}_P \subseteq \mathcal{E}\times\mathcal{P}\times\mathcal{L}$ is a set of property-facts (both can be represented as a simple list of triples). To simplify the notation, we assume that all inverse relations are also present in $\mathcal{F}_R$ -- that is, if $\langle e,r,e' \rangle$ is in $\mathcal{F}_R$, we also have $\langle e',r^{-1}, e \rangle$ in $\mathcal{F}_R$, effectively doubling the number of possible relations in the $\KB$.\footnote{This allows us to look at only one standard direction of facts and cover all possibilities -- see for example how it is used in the definition of \tt{compatible-neigbhors} in \eqref{eq:compatible}.}

\textbf{Problem: one-to-one alignment of instances between two knowledge bases.}  Given two knowledge bases $\KB_1$ and $\KB_2$ as well as a partial mapping between their corresponding relationships and properties, we want to output a 1-1 partial mapping $m$ from $\mathcal{E}_1$ to $\mathcal{E}_2$ which represents the semantically equivalent entities in the two knowledge bases (by partial mapping, we mean that the domain of $m$ does not have to be the whole of $\mathcal{E}_1$). %

\subsection{Possible approaches} \label{ssec:approaches}
Standard approaches for the ontology matching problem, such as RiMOM~\cite{li09RiMOM}, could be used to align small knowledge bases. However, they do not scale to millions of entities as needed for our task given that they usually consider all pairs of entities, suffering from a quadratic scaling cost. On the other hand, the related problem of identifying duplicate entities known as \emph{record linkage} or \emph{duplicate detection} in the database field, and \emph{co-reference resolution} in the natural langue processing field, do have scalable solutions~\cite{arasu09deduplication,gracia09largeScaleSenses}, though these do not exploit the 1-1 matching combinatorial structure present in our task, which reduces their accuracy. More specifically, they usually make independent decisions for different entities using some kind of similarity function, rather than exploiting the competition between different assignments for entities.
A notable exception is the work on \emph{collective} entity resolution by Bhattacharya and Getoor~\cite{getoor07relational}, solved using a greedy agglomerative clustering algorithm. The algorithm \ts{SiGMa} that we present in Section~\ref{sec:algorithm} can actually be seen as an efficient specialization of their work to the task of knowledge base alignment.

\begin{figure}
	\begin{center}
		\includegraphics[width=\columnwidth]{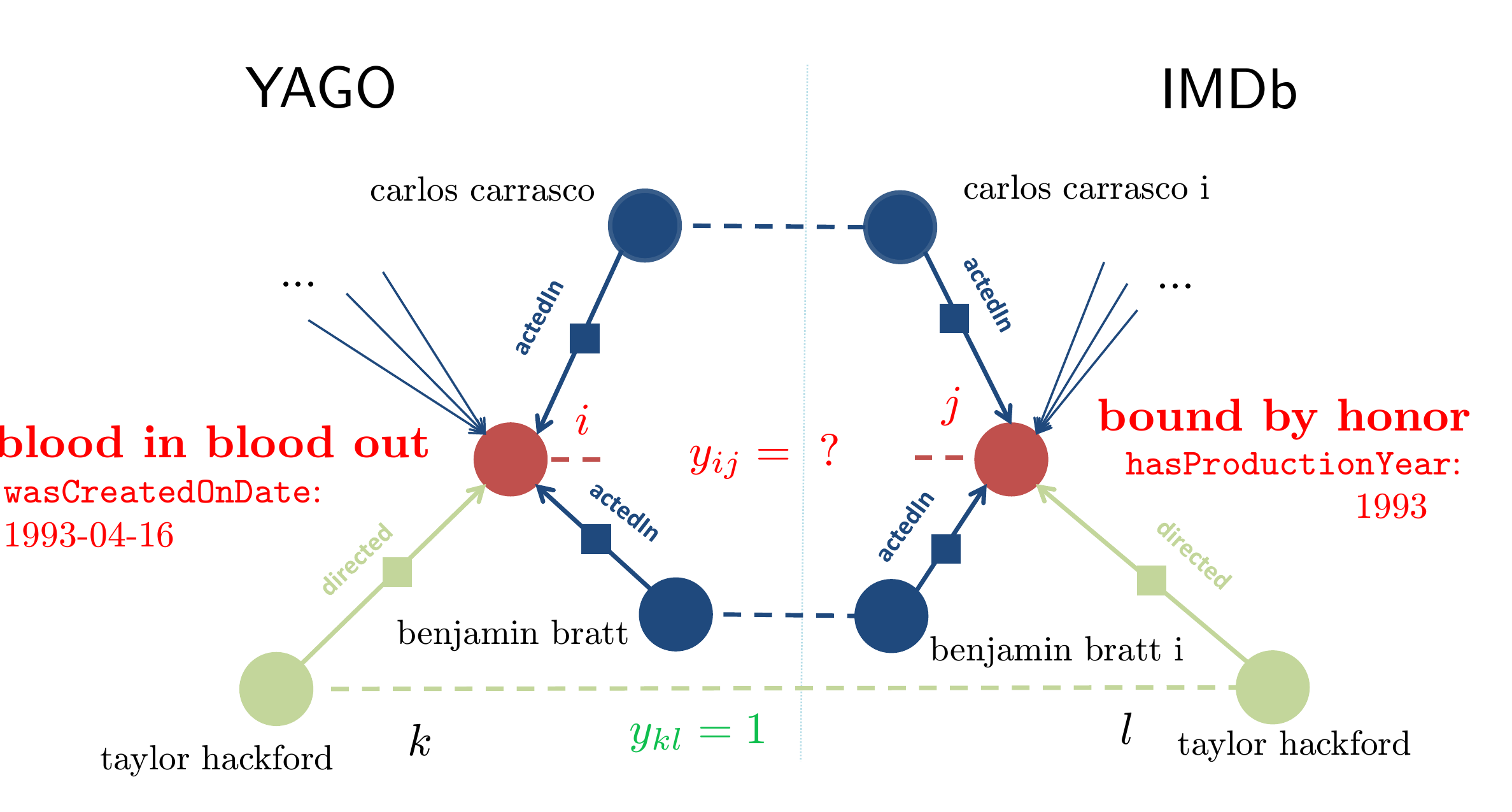}
	\end{center}
    \vspace*{-6mm}
	\caption{\textbf{Example of neighborhood to match in \ts{YAGO} and \ts{IMDb}.} \textnormal{Even though entities $i$ and $j$ have no words in common, the fact that several of their respective neighbors are matched together is a strong signal that $i$ and $j$ should be matched together. This is a real example from the dataset used in the experiments and \ts{SiGMa} was able to correctly match all these pairs ($i$ and $j$ are actually the same movie despite their different stored titles in each $\KB$).}} \label{fig:example}
\end{figure}

Another approach to alignment arises from the word alignment problem in natural language processing~\cite{och03comparison}, which has been formulated as a maximum weighted bipartite matching problem~\cite{taskar05matching} (thus exploiting the 1-1 matching structure). It also has been formulated as a quadratic assignment problem in~\cite{lacoste06qap}, which encourages neighbor entities in one graph to align to neighbor entities in the other graph, thus enabling alignment decisions to depend on each other --- see the caption of Figure~\ref{fig:example} for an example of this in our setup. The quadratic assignment formulation~\cite{lawler63qap}, which can be solved as an integer linear program, is NP-hard in general though, and these approaches were only used to align at most one hundred entities. In the algorithm \ts{SiGMa} that we propose, we are interested in exploiting both the 1-1 matching constraint, as well as building on previous decisions, like these word alignment approaches, but in a scalable manner which would handle millions of entities. \ts{SiGMa} does this by greedily optimizing the quadratic assignment objective, as we will describe in Section~\ref{ssec:greedy}. Finally, Suchanek et al.~\cite{suchanek12PARIS} recently proposed an ontology matching approach called \ts{PARIS} that they have succeeded to apply on the alignment of \ts{YAGO} to \ts{IMDb} as well, though the scalability of their approach is not as clear, as we will explain in Section~\ref{sec:related}. We will provide a detailed comparison with \ts{PARIS} in the experiments section.

\subsection{Design choices and assumptions}
Our main design choices result from our need for a fast algorithm for knowledge base alignment which scales to millions of entities. %
To this end we made the following assumptions:

\textbf{1-1 matching and uniqueness.} We assume that the true alignment between the two $\KB$s is a partial function which is mainly 1-1. %
If there are duplicate entities inside a $\KB$, \textsf{SiGMa} will only align one of the duplicates to the corresponding entity in the other $\KB$.

\textbf{Aligned relationships.} We assume that we are given a partial alignment between relationships and between properties of the $\KB$s.

\section{The SiGMa Algorithm} \label{sec:algorithm}
\subsection{Greedy optimization of a quadratic assignment objective} \label{ssec:greedy}
The \textsf{SiGMa} algorithm can be seen as the greedy optimization of an objective function which globally scores the suitability of a particular matching $m$ for a pair of given $\KB$s. This objective function will use two sources of information useful to choose matches: a similarity function between pairs of entities defined from their properties; and a graph neighborhood contribution making use of neighbor pairs being matched (see Figure~\ref{fig:example} for a motivation). Let us encode the matching $m : \mathcal{E}_1 \rightarrow \mathcal{E}_2 $ by a matrix $y$ with entries indexed by the entities in each $\KB$, with $y_{ij} = 1$ if $m(i) = j$, meaning that $i \in \mathcal{E}_1$ is matched to $j \in \mathcal{E}_2$, and $y_{ij} = 0$ otherwise. The space of possible 1-1 partial mappings is thus represented by the set of binary matrices: $\mathcal{M} \doteq \{y \in \{0,1\}^{\mathcal{E}_1 \times \mathcal{E}_2} : \sum_{l} y_{il} \leq 1 \; \forall i \in \mathcal{E}_1$ and $\sum_{k} y_{kj} \leq 1 \; \forall{j} \in \mathcal{E}_2\}$. We define the following quadratic objective function which globally scores the suitability of a matching $y$:
\begin{equation} \label{eq:obj}
    \begin{split}
    \texttt{obj}(y) & \doteq \sum_{(i,j) \in \mathcal{E}_1  \times \mathcal{E}_2} y_{ij} \left[ (1-\alpha) s_{ij} + \alpha g_{ij}(y) \right], \\
    & \textrm{where} \qquad g_{ij}(y) \doteq \sum_{(k,l) \in \mathcal{N}_{ij}} y_{kl} \, w_{ij,kl}.
    \end{split}
\end{equation}
The objective contains linear coefficients $s_{ij}$ which encode a similarity between entity $i$ and $j$, as well as quadratic coefficients $w_{ij,kl}$ which control the algorithm's tendency to match $i$ with $j$ given that $k$ was matched to $l$\footnote{In the rest of this paper, we will use the convention that $i$ and $k$ are always entities in $\KB_1$; whereas $j$ and $l$ are in $\KB_2$. $e$ could be in either $\KB$.}. $\mathcal{N}_{ij}$ is a local neighborhood around $(i,j)$ that we define later and which will depend on the graph information from the $\KB$s -- $g_{ij}(y)$ is basically counting (in a weighted fashion) the number of matched pairs $(k,l)$ which are in the neighborhood of $i$ and $j$. $\alpha \in [0,1]$ is a tradeoff parameter between the linear and quadratic contributions. Our approach is motivated by the maximization problem:
\begin{equation} \label{eq:opt}
\begin{aligned}
\max_y &\quad \texttt{obj}(y) \\
\textrm{s.t.} &\quad y \in \mathcal{M}, \quad \lVert y \rVert_1 \leq R,
\end{aligned}
\end{equation}
where the norm $\lVert y \rVert_1 \doteq \sum_{ij} y_{ij}$ represents the number of elements matched
and $R$ is an unknown upper-bound which represents the size of the best partial mapping which can be made from $\KB_1$ to $\KB_2$. We note that if the coefficients are all positive (as will be the case in our formulation -- we are only encoding similarities and not repulsions between entities), then the maximizer $y^*$ will have $\lVert y^* \rVert_1 = R$. Problem~\eqref{eq:opt} is thus related to one of the variations of the quadratic assignment problems, a well-known NP-complete problem in operational research~\cite{lawler63qap}\footnote{See Appendix~\ref{ap:qap} for the traditional description of the quadratic assignment problem and its relationship to our problem.}.
Even though one could approximate the solution to the combinatorial optimization~\eqref{eq:opt} using a linear program relaxation (see Lacoste-Julien et al.~\cite{lacoste06qap}), the number of variables is quadratic in the number of entities, and so is obviously not scalable. Our approach is instead to \emph{greedily optimize}~\eqref{eq:opt} by adding the match element $y_{ij}=1$ at each iteration which increases the objective the most and selected amongst a small set of possibilities. In other words, the high-level operational definition of the \textsf{SiGMa} algorithm is as follows:
\begin{enumerate}
    \item Start with an initial good quality partial match $y_0$.
    \item At each iteration $t$, augment the previous matching with a new matched pair by setting $y_{ij}=1$ for the $(i,j)$ which maximally increases $\texttt{obj}$, chosen amongst a small set $\mathcal{S}_t$ of reasonable candidates which preserve the feasibility of the new matching.
    \item Stop when the bound $\lVert y \rVert_1 = R$ is reached (and never undo previous decisions).
\end{enumerate}

Having outlined the general framework, in the remainder of this section we will describe methods for choosing the similarity coefficients $s_{ij}$ and $w_{ij,kl}$ so that they guide the algorithm towards good matchings (Section~\ref{sec:score}), the choice of neighbors, $\mathcal{N}_{ij}$, the choice of a candidate set $\mathcal{S}_t$, and the stopping criterion, $R$. These choices influence both the speed and accuracy of the algorithm.

\textbf{Compatible-neighbors.} $\mathcal{N}_{ij}$ should be chosen so as to respect the graph structure defined by the $\KB$ facts. Its contribution in the objective crucially encodes the fact that a neighbor $k$ of $i$ being matched to a `compatible' neighbor $l$ of $j$ should encourage $i$ to be matched to $j$ --- see the caption of Figure~\ref{fig:example} for an example. Here, compatibility means that they are related by the same relationship (they have the same color in Figure~\ref{fig:example}). Formally, we define:
\begin{multline} \label{eq:compatible}
    \mathcal{N}_{ij} = \textrm{\tt{compatible-neighbors}}(i,j) \doteq \\
    \textrm{\parbox{0.8\columnwidth}{\{ $(k,l)$ : $\langle i,r,k \rangle$ is in $\mathcal{F}_{R1}$ and $\langle j,s,l \rangle$ is in $\mathcal{F}_{R2}$ and relationship $r$ is matched to $s$\}.}}
\end{multline}
Note that a property of this neighborhood is that $(k,l) \in \mathcal{N}_{ij}$ iff $(i,j) \in \mathcal{N}_{kl}$, as we have that the relationship $r$ is matched to $s$ iff $r^{-1}$ is matched to $s^{-1}$ as well. This means that the increase in the objective obtained by adding $(i,j)$ to the current matching $y$ defines the following \emph{context dependent similarity score function} which is used to pick the next matched pair in the step 2 of the algorithm:
    \begin{multline} \label{eq:score}
       \texttt{score}(i,j; y)  =  (1-\alpha) s_{ij} + \alpha \, \delta g_{ij}(y)   \\
        \textrm{where }  \delta g_{ij}(y)  \doteq \sum_{(k,l) \in \mathcal{N}_{ij}} y_{kl} \, (w_{ij,kl} + w_{kl,ij}).
    \end{multline}

\textbf{Information propagation on the graph.} The \tt{compati\-ble-neighbors} concept that we just defined is one of the most crucial characteristics of \ts{SiGMa}. It allows the information of a new matched pair to propagate amongst its neighbors. It also defines a powerful heuristic to suggest new candidate pairs to include in a small set $\mathcal{S}_t$ of matches to choose from: after matching $i$ to $j$, \ts{SiGMa} adds all the pairs $(k,l)$ from \tt{compatible-neighbors}$(i,j)$ as new candidates. This yields a fire propagation analogy for the algorithm: starting from an initial matching (fire) -- it starts to match their neighbors, letting the fire propagate through the graph. If the graph in each $\KB$ is well-connected in a similar fashion, it can visit most nodes this way. This heuristic enables \ts{SiGMa} to avoid the potential quadratic number of pairs to consider by only focussing its attention on the neighborhoods of current matches.

\textbf{Stopping criterion.} \ts{SiGMa} terminates when the variation in the objective value, \texttt{score}$(i,j; y)$, of the latest added match $(i,j)$ falls below a threshold (or the queue becomes empty). The threshold in effect controls the precision / recall tradeoff of the algorithm. By ensuring that the $s_{ij}$ and $g_{ij}(y)$ terms are normalized between 0 and 1, we can standardize the scale of the threshold for different score functions. In our experiments, a threshold of 0.25 is observed to correlate well with a point at which the F-measure stops increasing and the precision is significantly decreasing.

\subsection{Algorithm and implementation} \label{ssec:implementation}
We present the pseudo-code for \ts{SiGMa} in Table~\ref{tab:alg}. We now elaborate on the algorithm design as well as its implementation aspects. We note that the \texttt{score} defined in~\eqref{eq:score} to greedily select the next matched pair is composed of a static term $s_{ij}$, which does not depend on the evolving matching $y$, and a dynamic term $\delta g_{ij} (y)$, which depends on $y$, though only through the local neighborhood $\mathcal{N}_{ij}$. We call the $\delta g_{ij}$ component of the score function the graph contribution -- its local dependence means that it can be updated efficiently after a new match has been added. We explain in more details the choice of similarity measures for these components in Section~\ref{sec:score}.

{\addtolength{\textfloatsep}{-0.2\textfloatsep}
\begin{table}
\fbox{\parbox[t]{\columnwidth}{
\begin{algorithmic}[1]
    \STATE Initialize matching $m=m_0$.
    \STATE Initialize priority queue $\mathcal{S}$ of suggested candidate pairs as $\mathcal{S}_0 \cup \left(\bigcup_{(i,j) \in m} \mathcal{N}_{ij} \right)$ -- the \tt{compatible-neigbhors} of pairs in $m$, with $\texttt{score}(i,j; m)$ as their key.
    \WHILE{priority queue $\mathcal{S}$ is not empty}
        \STATE Extract $\langle \textrm{\tt{score}},i,j\rangle$ from queue $\mathcal{S}$
        \STATE \textbf{if} $\tt{score} \leq \tt{threshold}$ \textbf{then stop}
        \IF{$i$ or $j$ is already matched to some entity} \STATE skip them and \textbf{continue} loop
        \ELSE \STATE Set $m(i)=j$. \\
            \COMMENT{We update candidate lists and scores:}
            \FOR{$(k,l)$ in $\mathcal{N}_{ij}$ and not already matched}
                \STATE Add $\langle \textrm{\tt{score}}(k,l; m),k,l\rangle$ to queue $\mathcal{S}$.
            \ENDFOR
        \ENDIF
    \ENDWHILE
\end{algorithmic}}}
\caption{\textsf{SiGMa} algorithm.} \label{tab:alg}
\end{table}
}

\paragraph*{Initial match structure $m_0$} The algorithm can take any initial matching seed assumed of good quality. In our current implementation, this is done by looking for entities with the same string representation (with minimal standardization such as removing capitalization and punctuation) with an \emph{unambiguous 1-1 match} -- that is, we do not include an exact matched pair when more than two entities have this same string representation, thereby increasing precision. %

\paragraph*{Increasing score function with local dependence} The score function has a component $s_{ij}$ which is static (fixed at the beginning of the algorithm) from the properties of entities such as their string representation, and a component $\delta g_{ij}(y)$ which is dynamic, looking at how many neighbors are correctly matched. The dynamic part can actually only increase when new neighbors are matched, and only the scores of neighbors can change when a new pair is matched.

\paragraph*{Optional static list of candidates $\mathcal{S}_0$} Optionally, we can initialize $\mathcal{S}$ with a static list $\mathcal{S}_0$ which only needs to be scored once as any score update will come from neighbors already covered by step 11 of the algorithm. $\mathcal{S}_0$ has the purpose to increase the possible exploration of the graph when another strong source of information (which is not from the graph) can be used. In our implementation, we use an inverted index built on words to efficiently suggest entities which have at least two words in common in their string representation as potential candidates.\footnote{To keep the number of suggestions manageable, we exclude a list of stop words built automatically from the 1,000 most frequent words of each $\KB$.} %

\paragraph*{Data-structures} We use a binary heap for the priority queue implementation---insertions will thus be $O(\log n)$ where $n$ is the size of the queue. Because the score function can only increase as we add new matches, we do not need to keep track of stale nodes in the priority queue in order to update their scores, yielding a significant speed-up.

\subsection{Score functions} \label{sec:score}
An important factor for any matching algorithm is the similarity function between pairs of elements to match. Designing good similarity functions has been the focus of much of the literature on record linkage, entity resolution, etc., and because \textsf{SiGMa} uses the score function in a modular fashion, \textsf{SiGMa} is free to use most of them for the term $s_{ij}$ as long as they can be computed efficiently. We provide in this section our implementation choices (which were motivated by simplicity), but we note that the algorithm can easily handle more powerful similarity measures. The generic score function used by \textsf{SiGMa} was given in~\eqref{eq:score}. In the current implementation, the static part $s_{ij}$ is defined through the \emph{properties} of entities only. The graph part $\delta g_{ij}(y)$ depends on the \emph{relationships} between entities (as this is what determines the graph), as well as the previous matching $y$. We also make sure that $s_{ij}$ and $g_{ij}$ stay normalized so that the score of different pairs are on the same scale.

\subsubsection{Static  similarity measure}
The static property similarity measure is further decomposed in two parts: we single out a contribution coming from the string representation property of entities (as it is such a strong signal for our datasets), and we consider the other properties together in a second term: %
\begin{equation} \label{eq:static}
s_{ij} = (1-\beta) \textrm{\tt{string}}(i,j) + \beta \textrm{\tt{prop}}(i,j),
\end{equation}
where $\beta \in [0,1]$ is a tradeoff coefficient between the two contributions set to 0.25 during the experiments.
\paragraph*{String similarity measure} For the string similarity measure, we primarily consider the number of words which two strings have in common, albeit weighted by their information content. In order to handle the varying lengths of strings, we use the Jaccard similarity coefficient between the sets of words, a metric often used in information retrieval and other data mining fields~\cite{hamers89Jaccard,getoor07relational}. The Jaccard similarity between set $A$ and $B$ is defined as $\textrm{Jaccard}(A,B) \doteq |A \cap B| / |A \cup B|$, which is a number between 0 and 1 and so is normalized as required. We also add a smoothing term in the denominator in order to favor longer strings with many words in common over very short strings. Finally, we use a \emph{weighted} Jaccard measure in order to capture the information that some words are more informative than others. In analogy to a commonly used feature in information retrieval, we use the IDF (inverse-document-frequency) weight for each word. The weight for word $v$ in $\KB_o$ is $w^o_v \doteq \log_{10} \frac{|\mathcal{E}_o|}{|E^o_v|}$, where $E^o_v \doteq \{e \in \mathcal{E}_o :$ $e$ has word $v$ in its string representation$\}$. Combining these elements, we get the following string similarity measure:
{\addtolength\abovedisplayskip{-4mm} %
\begin{equation} \label{eq:string}
\textrm{\tt{string}}(i,j) = \frac{\D \sum_{v \in \left(\mathcal{W}_i \cap  \mathcal{W}_j\right)} (w^1_v + w^2_v)}
    {\D \texttt{smoothing } + \sum_{v \in \mathcal{W}_i} w^1_v + \sum_{v' \in \mathcal{W}_j} w^2_{v'}},
\end{equation}
}where $\mathcal{W}_e$ is the set of words in the string representation of entity $e$ and \texttt{smoothing} is the scalar smoothing constant (we try different values in the experiments). Using unit weights and removing the smoothing term would recover the standard Jaccard coefficient between the two sets. As it operates on set of words, this measure is robust to word re-ordering, a frequently observed variation between strings representing the same entity in different knowledge bases. On the other hand, this measure is not robust to small typos or small changes of spelling of words. This problem could be addressed by using more involved string similarity measures such as \emph{approximate string matching}~\cite{chulman97wordMatching,stoilos05stringMetric}, which handles both word corruption as well as word reordering, though our current implementation only uses~\eqref{eq:string} for simplicity. We will explore the effect of different scoring functions in our experiments in Section~\ref{sec:parameters}.

\paragraph*{Property similarity measure} We recall that we assume that the user provided a partial matching between properties of both databases. This enables us to use them in a property similarity measure. In order to elegantly handle missing values of properties, varying number of property values present, etc., we also use a smoothed weighted Jaccard similarity measure between the sets of properties. The detailed formulation is given in Appendix~\ref{ap:property} for completeness, but we note that it can make use of a similarity measure between literals such a normalized distance on numbers (for dates, years etc.) or a string-edit distance on strings.

\subsubsection{Dynamic graph similarity measure}
We now introduce the part of the score function which enables \textsf{SiGMa} to build on previous decisions and exploit the relationship graph information. We need to determine $w_{ij,kl}$, the weight of the contribution of a neighboring matched pair $(k,l)$ for the score of the candidate pair $(i,j)$. The general idea of the graph score function is to count the number of compatible neighbors which are currently matched together for a pair of candidates (this is the $g_{ij}(y)$ contribution in~\eqref{eq:obj}). Going back at the example in Figure~\ref{fig:example}, there were three compatible matched pairs shown in the neighborhood of $i$ and $j$. We would like to normalize this count by dividing by the number of possible neighbors, and we would possibly want to weight each neighbor differently. We again use a smoothed weighted Jaccard measure to summarize this information, averaging the contribution from each $\KB$. This can be obtained by defining $w_{ij,kl}$ = $\gamma_{i} w_{ik} + \gamma_{j} w_{jl}$, where $\gamma_i$ and $\gamma_j$ are normalization factors specific to $i$ and $j$ in each database and $w_{ik}$ is the weight of the contribution of $k$ to $i$ in $\KB_1$ (and similarly for $w_{kl}$ in $\KB_2$). The graph contribution thus becomes:
\begin{equation} \label{eq:g_ij}
    g_{ij}(y) = \sum_{(k,l) \in \mathcal{N}_{ij}} y_{kl}  (\gamma_i w_{ik} + \gamma_j w_{jl}).
\end{equation}
So let $\mathcal{N}_{i}$ be the set of neighbors of entity $i$ in $\KB_1$, i.e. $\mathcal{N}_{i} \doteq \{ k : \exists r \textrm{ s.t. } (i,r,k) \in \mathcal{F}_{R1} \}$ (and similarly for $\mathcal{N}_j$). Then, remembering that $\sum_{k} y_{kl} \leq 1$ for a valid partial matching $y \in \mathcal{M}$, the following normalizations $\gamma_i$ and $\gamma_j$ will yield the average of two smoothed weighted Jaccard measures for $g_{ij}(y)$:
{\addtolength\abovedisplayskip{-3mm} %
\begin{equation} \label{eq:gamma_i}
    \gamma_i \doteq \frac{1}{2} \left(1 + \sum_{k \in \mathcal{N}_{i}} w_{ik} \right)^{-1} \gamma_j \doteq \frac{1}{2} \left(1 + \sum_{l \in \mathcal{N}_{j}} w_{jl} \right)^{-1}
\end{equation}
}We thus have $g_{ij}(y) \leq 1$ for $y \in \mathcal{M}$, keeping the contribution of each possible matched pair $(i,j)$ on the same scale in \texttt{obj} in~\eqref{eq:obj}.

The graph part of the score in~\eqref{eq:score} then takes the form:
\begin{equation} \label{eq:graphscore}
     \delta g_{ij}(y)  = \sum_{(k,l) \in \mathcal{N}_{ij}} y_{kl} \, ( \gamma_i w_{ik} + \gamma_j w_{jl} + \gamma_k w_{ki} + \gamma_l w_{lj}).
\end{equation}
The summation over the first two terms yields $g_{ij}(y)$ and so is bounded by $1$, but the summation over the last two terms could be greater than 1 in the case that $(i,j)$ is filling a `hole' in the graph (thus increasing the contribution of many neighbors $(k,l)$ in \texttt{obj} in~\eqref{eq:obj}). For example, suppose that $i$ has $n$ neighbors with degree 1 (i.e. they only have $i$ as neighbor); and the same thing for $j$, and that they are all matched pairwise --- Figure~\ref{fig:example} is an example of this with $n=3$ if we suppose that no other neighbors are present in the $\KB$. Suppose moreover that we use unit weights for $w_{ik}$ and $w_{jl}$. Then the normalization is $\gamma_k = 1/4$ for each $k \in \mathcal{N}_i$ (as they have degree 1); and similarly for $\gamma_l$. The contribution of the sum over the last two terms in~\eqref{eq:graphscore} is thus $n/2$ (whereas in this case $g_{ij}(y) = n/(n+1) \leq 1$).

\textbf{Neighbor weight $w_{ik}$.}
We finally need to specify the weight $w_{ik}$, which determines the strength of the contribution of the neighbor $k$ being correctly matched to the score of a suggested pair containing $i$. In our experiments, we consider both the constant weight $w_{ik} = 1$ and a weight $w_{ik}$ that varies inversely with the number of neighbors entity $k$ has where the relationship is of the same type as the one with entity $i$. The motivation for the latter is explained in Appendix~\ref{ap:weights}.

\section{Experiments} \label{sec:experiments}

\subsection{Setup} \label{ssec:setup}
We made a prototype implementation of \textsf{SiGMa} in Python\footnote{The code and datasets will be made available at \href{http://mlg.eng.cam.ac.uk/slacoste/sigma}{http://mlg.eng.cam.ac.uk/slacoste/sigma}.} and compared its performance on benchmark datasets as well as on large-scale knowledge bases. All experiments were run on a cluster node Hexacore Intel Xeon E5650 2.66GHz with 46GB of RAM running Linux. Each knowledge base is represented as two text files containing a list of triples of relationships-facts and property-facts. The input to \textsf{SiGMa} is a pair of such $\KB$s as well as a partial mapping between the relationships and properties of each $\KB$ which is used in the computation of the score in~\eqref{eq:score}, and the definition of \texttt{compatible-neighbors}~\eqref{eq:compatible}. The output of \textsf{SiGMa} is a list of matched pairs $(e_1,e_2)$ with their score information and the iteration number at which they were added to the solution. We evaluate the final alignment (after reaching the stopping threshold) by comparing it to ground truth using the standard metrics of precision, recall and F-measure on the number of \emph{entities} correctly matched.\footnote{Recall is defined in our setup as the number of correctly matched entities in $\KB_1$ divided by the number of entities with ground truth information in $\KB_1$. We note that recall is upper bounded by precision because our alignment is a 1-1 function.} The benchmark datasets are available together with corresponding ground truth data; for the large-scale knowledge bases, we built their ground truth using web url information as described in Section~\ref{sec:datasets}.

We found reasonable values for the parameters of \textsf{SiGMa} by
exploring its performance on the \ts{YAGO} to \ts{IMDb} pair (the
methodology is described in Section~\ref{sec:parameters}), and then
kept them fixed for all the other experimental comparisons
(Section~\ref{sec:exp1} and~\ref{sec:exp2}). This reflects the situation where one would like to apply \textsf{SiGMa} to a new dataset without ground truth or to minimize parameter adaptation. The standard parameters that we used in these experiments are given in Appendix~\ref{ap:params}.

\subsection{Datasets} \label{sec:datasets}

Our experiments were done both on several large-scale datasets and on
some standard benchmark datasets from the ontology alignment evaluation
initiative (OAEI) (Table~\ref{tab:all_stats}). We describe these datasets below.

\paragraph*{Large-scale datasets}
As mentioned throughout this paper so far, we used the dataset pair
\textsf{YAGO}-\textsf{IMDb}  as the main motivating example for
developing and testing \ts{SiGMa}. We also test \ts{SiGMa} on the pair
\ts{Freebase}-\textsf{IMDb}, for which we could obtain a sizable
ground truth. We describe here their construction. Both \textsf{YAGO}
and \ts{Freebase} are available as lists of triples from their respective
websites.\footnote{\textsf{YAGO2 core} was downloaded from:
  \href{http://www.mpi-inf.mpg.de/yago-naga/yago/downloads.html}{\small{http://www.mpi-inf.mpg.de/yago-naga/yago/downloads.html}}
  and \textsf{Freebase} from:
  \href{http://wiki.freebase.com/wiki/Data\_dumps}{\small{http://wiki.freebase.com/wiki/Data\_dumps}}.}
\textsf{IMDb}, on the other hand, is given as a list of text
files.\footnote{\href{http://www.imdb.com/interfaces\#plain}{http://www.imdb.com/interfaces\#plain}} There are
different files for different categories, e.g.: actors, producers,
etc. We use these categories to construct a list of triples containing
facts about movies and people. Because \ts{SiGMa} ignores
relationships and properties that are not matched between the $\KB$s,
we could reduce the size of \textsf{YAGO} and \ts{Freebase} by keeping
only those facts which had a 1-1 mapping with \ts{IMDb} as presented in Table~\ref{tab:largeKB}, and the entities appearing in these facts. To facilitate the comparison of \ts{SiGMa} with \ts{PARIS}, the authors of~\ts{PARIS} kindly provided us their own version of \ts{IMDb} that we will refer from now on as \textsf{IMDb\_PARIS} --- this version has actually a richer structure in terms of properties. We also kept in \textsf{YAGO} the relationships and properties which were aligned with those of \textsf{IMDb\_PARIS} (Table~\ref{tab:largeKB}). Table~\ref{tab:all_stats} presents the number of unique entities and relationship-facts included in the relevant reduced datasets. We constructed the ground truth for \textsf{YAGO}-\textsf{IMDb} by scraping the relevant Wikipedia pages of entities to extract their link to the corresponding \textsf{IMDb} page, which often appears in the `external links' section. We then obtained the entity name by scraping the corresponding \textsf{IMDb} page and matched it to our constructed database by using string matching (and some manual cleaning). We obtained 54K ground truth pairs this way. We used a similar process for \ts{Freebase}-\textsf{IMDb} by accessing the \textsf{IMDb} urls which were actually stored in the database. This yielded 293K pairs, probably one of the largest knowledge bases alignment ground truth sets to date.
\begin{table}
\begin{center}
\resizebox{1.0\columnwidth}{!}{
\begin{tabular}{@{}cccc@{}}
\hline
\textbf{YAGO} & \textbf{IMDb\_PARIS} & \textbf{IMDb}  &\textbf{Freebase}\\ \hline
 \multicolumn{4}{c}{\tt{Relations}} \\ \hline
actedIn & actedIn   & actedIn & actedIn\\
directed & directorOf  & directed & directed\\
produced & producerOf  & produced & produced\\
created & writerOf & composed & \\
wasBornIn & bornIn & &\\
diedIn & deceasedIn & &\\
capitalOf & locatedIn & &\\ \hline
 \multicolumn{4}{c}{\tt{Properties}} \\ \hline
hasLabel & hasLabel   & hasLabel & hasLabel\\
wasCreatedOnDate  &    & hasProductionYear & initialReleaseDate\\
wasBornOnDate & bornOn   &  & \\
diedOnDate & deceasedOn   &  & \\
hasGivenName  & firstName   &  & \\
hasFamilyName  & lastName   &  & \\
hasGender & gender   &  & \\
hasHeight  & hasHeight  &  & \\ \hline
\end{tabular}
}
\end{center}
\caption{Manually aligned movie related relationships and properties in large-scale $\KB$s.} \label{tab:largeKB}
\end{table}

\paragraph*{Benchmark datasets}

We also tested \ts{SiGMa} on three benchmark dataset pairs provided by
the ontology alignment evaluation initiative (OAEI), which allowed us
to compare the performance of \ts{SiGMa} to some previously published methods~\cite{li09RiMOM,hu11objectCoref}.
 From the OAEI 2009
 edition,\footnote{\href{http://oaei.ontologymatching.org/2009/instances/}{http://oaei.ontologymatching.org/2009/instances/}}
 we use the \textsf{Rexa}-\textsf{DBLP} instance matching benchmark
 from the domain of scientific publications.\footnote{We note that the
   smaller \ts{eprints} dataset also present in the benchmark was not
   suitable for 1-1 matchings as its ground truth had a large number
   of many-to-one matches.} \ts{Rexa} contains publications and
 authors as entities extracted from the search results of the \ts{Rexa} search
 server. \textsf{DBLP} is a version of the DBLP dataset
 listing publications from the computer science domain. The pair has
 one matched relationship, \texttt{author}, as well several matched
 properties such as \tt{year}, \tt{volume}, \tt{journal name},
 \tt{pages}, etc. Our goal was to align publications and authors. The other two datasets come from the Person-Restaurants (PR) task from the OAEI 2010 edition,\footnote{\href{http://oaei.ontologymatching.org/2010/im/index.html}{http://oaei.ontologymatching.org/2010/im/index.html}} containing data about people and restaurants. In particular, there are \textsf{person11}-\textsf{person12} pairs where the second entity is a copy of the first with one property field corrupted, and \textsf{restaurant1}-\textsf{restaurants2} pairs coming from two different online databases that were manually aligned. All datasets were downloaded from the corresponding OAEI webpages, with dataset sizes given in Table~\ref{tab:all_stats}.
\begin{table}
\centering
\resizebox{0.5\columnwidth}{!}{
\begin{tabular}{ccc}
\hline
\textbf{Dataset} & \textbf{\#facts} & \textbf{\#entities}\\ \hline
\textsf{YAGO}  & 442K   & 1.4M \\
\textsf{IMDb\_PARIS} & 20.9M  & 4.8M\\
\textsf{IMDb} & 9.3M  & 3.1M \\
\textsf{Freebase} & 1.5M & 474K  \\ \hline
\textsf{DBLP}  &  2.5M & 1.6M\\
\textsf{Rexa} & 12.6K & 14.7K\\
\textsf{person11} & 500 & 1000 \\
\textsf{person12} & 500 & 1000 \\
\textsf{restaurant1} & 113 & 339 \\
\textsf{restaurant2} & 752 & 2256 \\ \hline
\end{tabular}
}
\caption{Datasets statistics}\label{tab:all_stats}
\end{table}

\subsection{Exp. 1: Large-scale alignment} \label{sec:exp1}
In this experiment, we test the performance of \ts{SiGMa} on the three
pairs of large-scale $\KB$s and compare it with
\ts{PARIS}~\cite{suchanek12PARIS}, which is described in more details
in the related work Section~\ref{sec:related}. We also compare
\ts{SiGMa} and \ts{PARIS} with the simple baseline of doing the unambiguous exact string matching step described in Section~\ref{ssec:implementation} which is used to obtain an initial match $m_0$ (called \ts{Exact-string}). Table~\ref{tab:res_large} presents the results.
\begin{table}
\centering
\resizebox{\columnwidth}{!}{\begin{tabular}{l lrrrr c r}
\hline
\textbf{Dataset} & \textbf{System} & \textbf{Prec} & \textbf{Rec} & \textbf{F }& \textbf{GT size} & \textbf{\# pred.} & \textbf{Time} \\ \hline
\multirow{2}{*}{\ts{Freebase}-\ts{IMdb}} & \ts{SiGMa} & 99 & 95 & 97 & \multirow{2}{*}{255k} & 366k & 90 min\\
 & \ts{Exact-string} & 99 & 70 & 82 &   & 244k & 1 min\\ \hline
\multirow{2}{*}{\ts{YAGO}-\ts{IMDb}} & \ts{SiGMa} & 98 & 93 & 95 &  \multirow{2}{*}{54k} & 188k & 50 min\\
     & \ts{Exact-string}& 99 & 57 & 72 &   & 162k & 1 min \\ \hline
\multirow{3}{*}{\specialcell{l}{\ts{YAGO}-\ts{IMDb\_PARIS} \\ (new ground truth) } } & \ts{SiGMa} & 98 & 96 & 97 &  & 237k & 70 min\\
 & PARIS & 97 & 96 & 97 & 57k & 702k & 3100 min\\
         & \ts{Exact-string} & 99 & 56 & 72 &   & 202k & 1 min \\ \hline
\multirow{3}{*}{\specialcell{l}{\ts{YAGO}-\ts{IMDb\_PARIS} \\ (ground truth from~\protect\cite{suchanek12PARIS}) }} & \ts{SiGMa} & 98 & 84 & 91 &  & 237k & 70 min\\
 & PARIS & 94 & 90 & 92 & 11k & 702k &  3100 min\\
 & \ts{Exact-string} & 99 & 61 & 75 &  & 202k &  1 min \\ \hline
\end{tabular}
}
\caption{Exp. 1: Results (precision, recall, F-measure) on large-scale datasets for \ts{SiGMa} in comparison to a simple exact-matching phase on strings as well as \ts{PARIS}~\protect\cite{suchanek12PARIS}. \textnormal{The `GT Size' column gives the number entities with ground truth information. Time is total running time, including loading the dataset (quoted from~\protect\cite{suchanek12PARIS} for \ts{PARIS}).}}\label{tab:res_large}
\end{table}
Despite its simple greedy nature which never goes back to correct a mistake, \ts{SiGMa} obtains an impressive F-measure above 90\% for all datasets, significantly improving over the \ts{Exact-string} baseline. We tried running \ts{PARIS}~\cite{suchanek12PARIS} on a smaller subset of \ts{YAGO}-\ts{IMDb}, using the code available from its author's website. It did not complete its first iteration after a week of computation and so we halted it (we did not have the SSD drive which seems crucial to reasonable running times). The results for \ts{PARIS} in Table~\ref{tab:res_large} are thus computed using the prediction files provided to us by its authors on the \ts{YAGO}-\ts{IMDb\_PARIS} dataset. In order to better relate the \ts{YAGO}-\ts{IMDb\_PARIS} results with the \ts{YAGO}-\ts{IMDb} ones, we also constructed a larger ground truth reference on \ts{YAGO}-\ts{IMDb\_PARIS} by using the same process as described in Section~\ref{sec:datasets}. On both ground truth evaluations, \ts{SiGMa} obtains a similar F-measure as \ts{PARIS}, but in 50x less time. On the other hand, we note that \ts{PARIS} is solving the more general problem of instances and schema alignment, and was not provided any manual alignment between relationships. The large difference of recall between \ts{PARIS} and \ts{SiGMa} on the ground truth from~\cite{suchanek12PARIS} can be explained by the fact that more than a third of its entities had no neighbor; whereas the process used to construct the new larger ground truth included only entities participating in movie facts and thus having at least one neighbor. The recall of \ts{SiGMa} actually increases for entities with increasing number of neighbors (going from 68\% for entities in the ground truth from~\cite{suchanek12PARIS} with 0 neighbor to 97\% for entities with 5+ neighbors).

About 2\% of the predicted matched pairs from \ts{SiGMa} on \ts{YAGO}-\ts{IMDb} have no word in common and thus zero string similarity -- difficult pairs to match without any graph information. Examples of these pairs came from spelling variations of names, movie titles in different languages, foreign characters in names which are not handled uniformly or multiple titles for movies (such as the `Blood In, Blood Out' example of Figure~\ref{fig:example}).

\textbf{Error analysis.} Examining the few errors made by \ts{SiGMa}, we observed the following types of matching errors: 1) errors in the ground truth (either coming from the scraping scheme used; or from Wikipedia (\ts{YAGO}) which had incorrect information); 2) having multiple very similar entities (e.g. mistaking the `making of' of the movie vs. the movie itself); 3) pair of entities which shared exactly the same neighbors (e.g. two different movies with exactly the same actors) but without other discriminating information. Finally, we note that going through the predictions of \ts{SiGMa} that had a low property score revealed a significant number of errors in the databases (e.g. wildly inconsistent birth dates for people), indicating that \ts{SiGMa} could be used to highlight data inconsistencies between databases.

\subsection{Exp. 2: Benchmark comparisons} \label{sec:exp2}
In this experiment, we test the performance of \ts{SiGMa} on the three benchmark datasets and compare them with the best published results so far that we are aware of: \ts{PARIS}~\cite{suchanek12PARIS} for the Person-Restaurants datasets (which compared favorably over ObjectCoref~\cite{hu11objectCoref}); and \ts{RiMoM}~\cite{li09RiMOM} for \ts{Rexa-DBPL}. Table~\ref{tab:res_bench} presents the results. We also include the results for \ts{Exact-string} as a simple baseline as well as \ts{SiGMa-linear}, which is the \ts{SiGMa} algorithm without using the graph information at all\footnote{\ts{SiGMa-linear} is not using the graph score component ($\alpha$ is set to 0) and is only using the inverted index $\mathcal{S}_0$ to suggest candidates -- not the neighbors in $\mathcal{N}_{ij}$.}, to give an idea of how important the graph information is in these cases.

Interestingly, \ts{SiGMa} significantly improved the previous results without needing any parameter tweaking. The Person-Restaurants datasets did not have a rich relationship structure to exploit: each entity (a person or a restaurant) was linked to exactly one another in a 1-1 bipartite fashion (their address). This is perhaps why \ts{SiGMa-linear} is surprisingly able to \emph{perfectly} match both the Person and Restaurants datasets. Analyzing the errors made by \ts{SiGMa}, we noticed that they were due to a violation of the assumption that each entity is unique in each $\KB$: the same address is represented as different entities in \ts{Restaurant2}, and \ts{SiGMa} greedily matched the one which was not linked to another restaurant in \ts{Restaurant2}, thus reducing the graph score for the correct match. \ts{SiGMa-linear} couldn't suffer from this problem, and thus obtained a perfect matching.

The \ts{Rexa-DBLP} dataset has a more interesting relationship structure which is not just 1-1: papers have multiple authors and authors have written multiple papers, enabling the fire propagation algorithm to explore more possibilities. However, it appears that a purely string based algorithm can already do quite well on this dataset --- \ts{Exact-string} obtains a 89\% F-measure, already significantly improving the previously best published results (\ts{RiMOM} at 76\% F-measure). \ts{SiGMa-linear} improves this to 91\%, and finally using the graph structure helps to improve this to 94\%. This benchmark which has a medium size also highlights the nice scalability of \ts{SiGMa}: despite using the interpreted language Python, our implementation runs in less than 10 minutes on this dataset, which can be compared to \ts{RiMOM} taking 36 hours on a 8-core server in 2009.
\begin{table}
\centering
\resizebox{\columnwidth}{!}{
\begin{tabular}{llrrrr}
\hline
\textbf{Dataset} & \textbf{System} & \textbf{Prec} & \textbf{Rec} & \textbf{F }& \textbf{GT size}\\ \hline
Person & \ts{SiGMa} & 100 & 100 & 100 & \multirow{2}{*}{500} \\
 & \ts{PARIS} & 100 & 100 & 100 & \\ \hline
Restaurant & \ts{SiGMa}-\ts{linear} & 100 & 100 & 100 & \multirow{4}{*}{89} \\ %
 & \ts{SiGMa} & 98 & 96 & 97 &  \\
 & \ts{PARIS} & 95 & 88 & 91 & \\
 & \ts{Exact-string} & 100 & 75 & 86 & \\ \hline
\ts{Rexa}-\ts{DBLP} & \ts{SiGMa} & 97 & 90 & 94 & \multirow{4}{*}{1464} \\
 & \ts{SiGMa-linear} & 96 & 86 & 91 & \\
 & \ts{Exact-string} & 98 & 81 & 89 & \\
 & \ts{RiMOM} & 80 & 72 & 76 & \\ \hline
\end{tabular}
}
\caption{Exp. 2: Results on the benchmark datasets for \ts{SiGMa}, compared with \ts{PARIS}~\protect\cite{suchanek12PARIS} and \ts{RiMOM}~\protect\cite{li09RiMOM}. \ts{SiGMa-linear} and \ts{Exact-string} are also included on the interesting datasets as further comparison points.}\label{tab:res_bench}
\end{table}

\subsection{Parameter experiments} \label{sec:parameters}
In this section, we explore the role of different configurations for \ts{SiGMa} on the \ts{YAGO}-\ts{IMDb} pair, as well as determine which parameters to use for the other experiments. We recall that \ts{SiGMa} with the final parameters (described in Appendix~\ref{ap:params}) yields a 95\% F-measure on this dataset (second section of Table~\ref{tab:res_large}). Experiments 5 and 6 which explore the optimal weighting schemes as well as the correct stopping threshold are described for completeness in Appendix~\ref{ap:param_exp}.

\subsubsection{Exp. 3: Score components}
\begin{figure}
	\begin{center}
    \vspace*{-7mm}
    \includegraphics[width=\columnwidth]{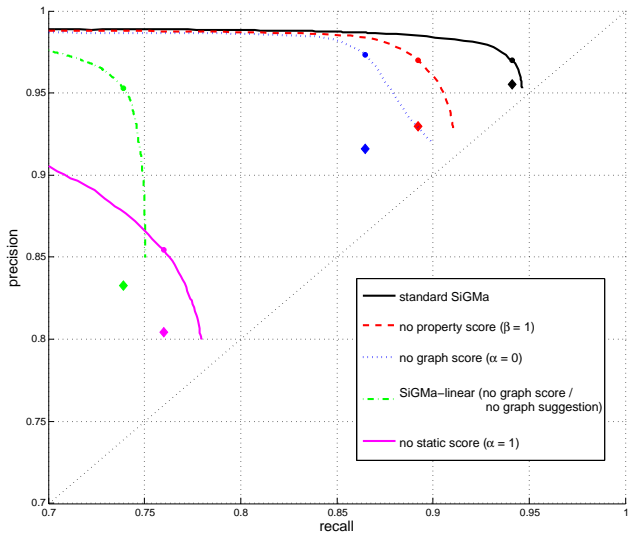}
    \vspace*{-8mm}
    \end{center}
	\caption{\textbf{Exp. 3: Precision/Recall curves for \ts{SiGMa} on \ts{YAGO}-\ts{IMDb} with different scoring configurations.} \textnormal{The filled circles indicate the maximum F-measure position on each curve, with the corresponding diamond giving the F-measure value at this recall point.} \label{fig:test_scores} }
\end{figure}

In this experiment, we explore the importance of each part of the score function by running \ts{SiGMa} with some parts turned off (which can be done by setting the $\alpha$ and $\beta$ tradeoffs to 0 or 1). The resulting precision / recall curves are plotted in Figure~\ref{fig:test_scores}a. We can observe that turning off the static part of the score (string and property) has the biggest effect, decreasing the maximum F-measure from 95\% to about 80\% (to be contrasted with the 72\% F-measure for \ts{Exact-string} as shown in Table~\ref{tab:res_large}). By comparing \ts{SiGMa} with \textsf{SiGMa-linear}, we see that including the graph information moves the F-measure from a bit below 85\% to over 95\%, a significant gain, indicating that the graph structure is more important on this dataset than the OAEI benchmark datasets.

\subsubsection{Exp. 4: Matching seed}
In this experiment, we tested how important the size of the matching seed $m_0$ is for the performance of \ts{SiGMa}. We report the following notable results. We ran \ts{SiGMa} with no exact seed matching at all: we initialized it with a random exact match pair and let it explore the graph greedily (with the inverted index still making suggestions). This obtained an even better score than the standard setup: 99\% of precision, 94\% recall and 96\% F-measure, demonstrating that \emph{a good initial seed is actually not needed for this setup}. If we do not use the inverted index but initialize \ts{SiGMa} with the top 5\% of the exact match sorted by their score in the context of the whole exact match, the performance drops a little, but \ts{SiGMa} is still able to explore a large part of the graph: it obtains 99\% / 87\% / 92\% of precision/recall/F-measure, illustrating the power of the graph information for this dataset.

\section{Related work} \label{sec:related}
We contrast here \ts{SiGMa} with the work already mentioned in Section~\ref{ssec:approaches} and provide further links. In the ontology matching literature, the only approach which was applied to datasets of the size that we considered in this paper is the recently proposed \ts{PARIS}~\cite{suchanek12PARIS}, which solves the more general problem of matching instances, relationships and classes. The \ts{PARIS} framework defines a normalized score between pairs of instances to match representing how likely they should be matched,\footnote{The authors call these `marginal probabilities' as they were motivated from probabilistic arguments, but these do not sum to one.} and which depends on the matching scores of their compatible neighbors. The final scores are obtained by first initializing (and fixing) the scores on pairs of literals, and then propagating the updates through the relationship graph using a fixed point iteration, yielding an analogous fire propagation of information as \ts{SiGMa}, though it works with soft [0-1]-valued assignment whereas \ts{SiGMa} works with hard \{0,1\}-valued ones. The authors handle the scalability issue of maintaining scores for all pairs by using a sparse representation with various pruning heuristics (in particular, keeping only the maximal assignment for each entity at each step, thus making the same 1-1 assumption that we did).  An advantage of \ts{PARIS} over \ts{SiGMa} is that it is able to include property values in its neighborhood graph (it uses soft-assignments between them) whereas \ts{SiGMa} only uses relationships given that a 1-1 matching of property values is not appropriate. We conjecture that this could explain the higher recall that \ts{PARIS} obtained on entities which had no relationship neighbors on the \ts{YAGO}-\ts{PARIS\_IMDB} dataset. On the other hand, \ts{PARIS} was limited to use a 0-1 similarity measure between property values for the large-scale experiments in~\cite{suchanek12PARIS}, as it is unclear how one could apply the same sparsity optimization in a scalable fashion with more involved similarity measures (such as the IDF one that \ts{SiGMa} is using). The use of a 0-1 similarity measure on strings could explain the lower performance of \ts{PARIS} on the Restaurants dataset in comparison to \ts{SiGMa}. We stress that \ts{SiGMa} is able in contrast to use sophisticated similarity measures in a scalable fashion, and had a 50x speed improvement over \ts{PARIS} on the large-scale datasets.

The \ts{SiGMa} algorithm is related to the collective entity resolution approach of Bhattacharya and Getoor~\cite{getoor07relational}, which proposed a greedy agglomerative clustering algorithm to cluster entities based on previous decisions. Their approach could handle constraints on the clustering, including a $1-1$ matching constraint in theory, though it was not implemented. A scalable solution for collective entity resolution was proposed recently in~\cite{restogi11largeEM}, by treating the sophisticated machine learning approaches to entity resolution as black boxes (see references therein), but running them on small neighborhoods and combining their output using a message-passing scheme. They do not consider exploiting a $1-1$ matching constraint though, as most entity resolution or record linkage work.

The idea to propagate information on a relationship graph has been used in several other approaches for ontology matching~\cite{hu05GMO,mao07network}, though none were scalable for the size of knowledge bases that we considered. An analogous `fire propagation' algorithm has been used to align social network graphs in~\cite{narayanan11deanonymization}, though with a very different objective function (they define weights in each graphs and want to align edges which has similar weights). The heuristic of propagating information on a relationship graph is related to a well-known heuristic for solving Constraint Satisfactions Problems known as constraint propagation~\cite{bessiere2006constraint}. Ehrig and Staab~\cite{ehrig04QOM} mentioned several heuristics to reduce the number of candidates to consider in ontology alignment, including a similar one to \texttt{compatible-neighbors}, though they tested their approach only on a few hundred instances. Finally, we mention that Peralta~\cite{peralta07movieMatching} aligned the movie database MovieLens to IMDb through a combination of steps of manual cleaning with some automation. \ts{SiGMa} could be considered as an alternative which does not require manual intervention apart specifying the score function to use.

\section{Conclusion}
We have presented \ts{SiGMa}, a simple and scalable algorithm for the alignment of large-scale knowledge bases. Despite making greedy decisions and never backtracking to correct decisions, \ts{SiGMa} obtained a higher F-measure than the previously best published results on the OAEI benchmark datasets, and matched the performance of the more involved algorithm~\ts{PARIS} while being 50x faster on large-scale knowledge bases of millions of entities. Our experiments indicate that \ts{SiGMa} can obtain good performance over a range of datasets with the same parameter setting. On the other hand, \ts{SiGMa} is easily extensible to more powerful scoring functions between entities, as long as they can be efficiently computed.

Some apparent limitations of \ts{SiGMa} are a) that it cannot correct previous mistakes and b) cannot handle alignments other than 1-1. Addressing these in a scalable fashion which preserves high accuracy are open questions for future work. We note though that the non-corrective nature of the algorithm didn't seem to be an issue in our experiments. Moreover, pre-processing each knowledge base with a de-duplication method can help make the 1-1 assumption more reasonable, which is a powerful feature to exploit in an alignment algorithm. Another interesting direction for future work would be to use machine learning methods to learn the parameters of more powerful scoring function. In particular, the `learning to rank' model seems suitable to learn a score function which would rank the correctly labeled matched pairs above the other ones. The current level of performance of \ts{SiGMa} already makes it suitable though as a powerful generic alignment tool for knowledge bases and hence takes us closer to the vision of Linked Open Data and the Semantic Web.

\textbf{Acknowledgments:} We thank Fabian Suchanek and Pierre Senellart for sharing their code and answering our questions about \ts{PARIS}. We thank Guillaume Obozinski for helpful discussions. This research was supported by a grant from Microsoft Research Ltd. and a Research in Paris fellowship.

\bibliographystyle{abbrv}
\small{\bibliography{sigma}}

\begin{thebibliography}{10}

\bibitem{arasu09deduplication}
A.~Arasu, C.~R\'{e}, and D.~Suciu.
\newblock Large-scale deduplication with constraints using dedupalog.
\newblock In {\em Proc. ICDE}, 2009.

\bibitem{bessiere2006constraint}
C.~Bessiere.
\newblock Constraint propagation.
\newblock {\em Foundations of Artificial Intelligence}, 2:29--83, 2006.

\bibitem{getoor07relational}
I.~Bhattacharya and L.~Getoor.
\newblock Collective entity resolution in relational data.
\newblock {\em ACM TKDD}, 1(1), 2007.

\bibitem{lee08WWW}
C.~Bizer, T.~Heath, K.~Idehen, and T.~Berners-Lee.
\newblock Linked data on the web ({LDOW}2008).
\newblock In {\em Proc. WWW}, 2008.

\bibitem{castano08instanceMatching}
S.~Castano, A.~Ferrara, D.~Lorusso, and S.~Montanelli.
\newblock On the ontology instance matching problem.
\newblock In {\em Proc. DEXA}, 2008.

\bibitem{choi06survey}
N.~Choi, I.-Y. Song, and H.~Han.
\newblock A survey on ontology mapping.
\newblock {\em SIGMOD Rec.}, 35:34--41, 2006.

\bibitem{ehrig04QOM}
M.~Ehrig and S.~Staab.
\newblock {QOM - Quick Ontology Mapping}.
\newblock In {\em ISWC}, 2004.

\bibitem{euzenat07om-book}
J.~Euzenat and P.~Shvaiko.
\newblock {\em Ontology matching}.
\newblock Springer-Verlag, 2007.

\bibitem{wordnet}
C.~Fellbaum.
\newblock {\em Word{N}et: An Electronic Lexical Database}.
\newblock Bradford Books, 1998.

\bibitem{chulman97wordMatching}
J.~C. French, A.~L. Powell, and E.~Schulman.
\newblock Applications of approximate word matching in information retrieval.

\bibitem{gracia09largeScaleSenses}
J.~Gracia, M.~d'Aquin, and E.~Mena.
\newblock Large scale integration of senses for the semantic web.
\newblock In {\em Proc. WWW}, 2009.

\bibitem{hamers89Jaccard}
L.~Hamers, Y.~Hemeryck, G.~Herweyers, M.~Janssen, H.~Keters, R.~Rousseau, and
  A.~Vanhoutte.
\newblock Similarity measures in scientometric research: The {J}accard index
  versus {S}alton's cosine formula.
\newblock {\em Information Processing \& Management}, 25(3), 1989.

\bibitem{hu11objectCoref}
W.~Hu, J.~Chen, and Y.~Qu.
\newblock A self-training approach for resolving object coreference on the
  semantic web.
\newblock In {\em Proc. WWW}, 2011.

\bibitem{hu05GMO}
W.~Hu, N.~Jian, Y.~Qu, and Y.~Wang.
\newblock {GMO}: A graph matching for ontologies.
\newblock In {\em K-Cap Workshop in Integrating Ontologies}, 2005.

\bibitem{kalfoglou03om-state-of-the-art}
Y.~Kalfoglou and M.~Schorlemmer.
\newblock Ontology mapping: the state of the art.
\newblock {\em Knowl. Eng. Rev.}, 18:1--31, 2003.

\bibitem{lacoste06qap}
S.~Lacoste-Julien, B.~Taskar, D.~Klein, and M.~I. Jordan.
\newblock Word alignment via quadratic assignment.
\newblock In {\em Proc. HLT-NAACL}, 2006.

\bibitem{lawler63qap}
E.~L. Lawler.
\newblock The quadratic assignment problem.
\newblock {\em Management Science}, 9(4):586--599, 1963.

\bibitem{li09RiMOM}
J.~Li, J.~Tang, Y.~Li, and Q.~Luo.
\newblock Rimom: A dynamic multistrategy ontology alignment framework.
\newblock {\em IEEE Trans. on Knowl. and Data Eng.}, 21:1218--1232, 2009.

\bibitem{mao07network}
M.~Mao.
\newblock Ontology mapping: An information retrieval and interactive activation
  network based approach.
\newblock In {\em ISWC/ASWC}, 2007.

\bibitem{narayanan11deanonymization}
A.~Narayanan, E.~Shi, and B.~I.~P. Rubinstein.
\newblock Link prediction by de-anonymization: How we won the {K}aggle social
  network challenge.
\newblock In {\em IJCNN}, 2011.

\bibitem{och03comparison}
F.~J. Och and H.~Ney.
\newblock A systematic comparison of various statistical alignment models.
\newblock {\em Comput. Linguist.}, 29:19--51, 2003.

\bibitem{peralta07movieMatching}
V.~Peralta.
\newblock {Matching of MovieLens and IMDb Movie Titles}.
\newblock Technical report, APDM project, Laboratoire PRiSM, Universit\'{e} de
  Versailles, 2007.

\bibitem{restogi11largeEM}
V.~Rastogi, N.~Dalvi, and M.~Garofalakis.
\newblock Large-scale collective entity matching.
\newblock {\em Proc. VLDB Endow.}, 4:208--218, 2007.

\bibitem{shvaiko08challenges}
P.~Shvaiko and J.~Euzenat.
\newblock Ten challenges for ontology matching.
\newblock In {\em Proc. ODBASE}, 2008.

\bibitem{stoilos05stringMetric}
G.~Stoilos, G.~B. Stamou, and S.~D. Kollias.
\newblock A string metric for ontology alignment.
\newblock In {\em Proc. ISWC}, pages 624--637.

\bibitem{suchanek12PARIS}
F.~M. Suchanek, S.~Abiteboul, and P.~Senellart.
\newblock {PARIS}: Probabilistic alignment of relations, instances, and schema.
\newblock {\em PVLDB}, 5(3):157--168, 2011.

\bibitem{suchanek2007WWW}
F.~M. Suchanek, G.~Kasneci, and G.~Weikum.
\newblock {Yago: A Core of Semantic Knowledge}.
\newblock In {\em Proc. WWW}, 2007.

\bibitem{taskar05matching}
B.~Taskar, S.~Lacoste-Julien, and D.~Klein.
\newblock A discriminative matching approach to word alignment.
\newblock In {\em Proc. EMNLP}, 2005.

\bibitem{RDFs}
{W3C}.
\newblock {RDF} {P}rimer ({W3C} {R}ecommendation 2004-02-10).

\end{thebibliography}

\clearpage

\begin{appendix}

\section{Property similarity measure} \label{ap:property}
We describe here the property similarity measure used in our implementation. We use a smoothed weighted Jaccard similarity measure between the sets of properties defined as follows. Suppose that $e_1$ has properties $p_1, p_2, \ldots, p_{n_1}$ with respective literal values $v_1, v_2, \ldots, v_{n_1}$, and that $e_2$ has properties $q_1, q_2, \ldots, q_{n_2}$ with respective literal values $l_1, l_2, \ldots, l_{n_2}$. In analogy to the string similarity measure, we will also associate IDF weights to the possible property values $w^o_{p,v} \doteq \log_{10} \frac{N^o_p}{|E^o_{p,v}|}$ where $E^o_{p,v} \doteq \{e \in \mathcal{E}_o :$ $e$ has literal $v$  for property $p\}$ and $N^o_p$ is the total number of entities in knowledge base $o$ which have a value for property $p$. We then define the following property similarity measure:
\begin{equation} \label{eq:property}
\mathtt{prop}(i,j) = \frac{\D \sum_{(a,b) \in M_{12} } (w^1_{p_a,v_a} + w^2_{q_b,l_b}) \,\mathrm{Sim}_{p_a,q_b}(v_a,l_b)}
    {\D 2 + \sum_{a=1}^{n_1} w^1_{p_a,v_a} + \sum_{b=1}^{n_2} w^2_{q_b,l_b}}.
\end{equation}
where $M_{12}$ represents the property alignment: $M_{12} \doteq \{(a,b) : p_a \textrm{ is matched to } q_b\}$. $\mathrm{Sim}_{p_a,q_b}(v_a,l_b)$ is a $[0,1]$-valued similarity measure between literals; it could be a normalized distance on numbers (for dates, years, etc.), a string-edit distance on strings, etc. %

\section{Graph neighbor weight} \label{ap:weights}
\begin{figure}
	\begin{center}
		\includegraphics[width=0.5\columnwidth]{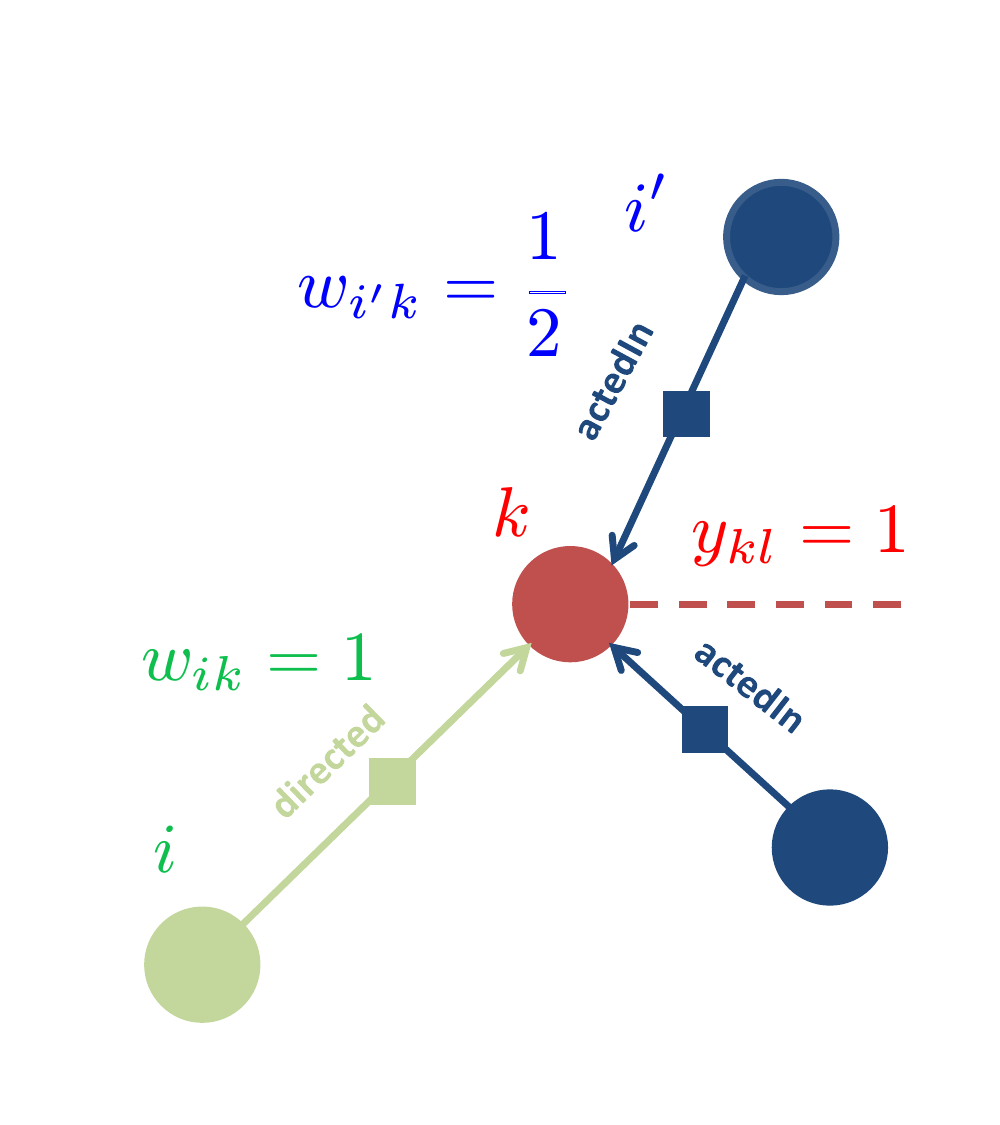}
	\end{center}
    \vspace*{-10pt}
	\caption{ Graph weight illustration. \normalfont{The contribution of the movie match $y_{kl} = 1$ should be weighted more for the candidate match pairing the only director $i$ of $k$ with a director of movie $l$ ($w_{ik} = 1$) as compared to the candidate match pairing one of the many actors $i'$ of $k$ with an actor of the movie $l$ ($w_{i'k} = 1/2$ for two actors in movie $k$). This weighting scheme can also be thought of ensuring that the contribution of the match $(k,l)$ spreads uniformly amongst all its neighbors with one unit of influence per relationship type in each $\KB$ separately.}} \label{fig:graph_weight}
\end{figure}

We recall that the the graph weight $w_{ik}$ determines the strength of the contribution of the neighbor $k$ being correctly matched to the score of a suggested pair containing $i$. In our experiments, we consider both the constant weight $w_{ik} = 1$ and a weight $w_{ik}$ that varies inversely with the number of neighbors entity $k$ has where the relationship is of the same type as the one with entity $i$. To motivate the latter, we go back again to our running example of Figure~\ref{fig:example}, but switching the role of $i$ and $k$ as we need to look at the neighbors of $k$ -- this is illustrated in Figure~\ref{fig:graph_weight} and explained in its caption. In case there are multiple different relationships linking the same pair $i$ to $k$, we take the maximum of the weights over these (i.e. we pick the most informative information to weight it). So formally, we have:
 \begin{equation} \label{eq:w_ik}
    w_{ik} \doteq \max_{r \textrm{ s.t. } (i,r,k) \in \mathcal{F}_R} | \{i' : (i',r,k) \in \mathcal{F}_R | ^{-1}  .
\end{equation}

We also point out that the normalization of $g_{ij}(y)$ in~\eqref{eq:g_ij} is made over each $\KB$ independently, in contrast with the \texttt{string} and \texttt{prop} similarity measures~\eqref{eq:string} and~\eqref{eq:property} which are normalized in both $\KB$ jointly. The motivation for this is that the neighborhood size in \ts{YAGO} and \ts{IMDb} are overly asymmetric (there is much more information about each movie in \ts{IMDb}). The separate normalization means that as long as most of a neighborhood in \emph{one} $\KB$ is correctly aligned, the graph score will be high. The information about strings and properties is more symmetric in the $\KB$ pairs that we consider, so a joint normalization seems reasonable in this case.

\section{Quadratic assignment problem} \label{ap:qap}
The quadratic assignment problem is traditionally defined as finding a bijection between $R$ facilities and $R$ locations which \emph{minimizes} the expected cost of transport between the facilities. Given that facilities $i$ and $k$ are assigned to locations $j$ and $l$ respectively, the cost of transport between facility $i$ and $k$ is $w_{ij,kl} = n_{ik} c_{jl}$, where $n_{ik}$ is the expected number of units to ship between facilities $i$ and $k$, and $c_{jl}$ is the expected cost of shipment between locations $j$ and $l$ (depending on their distance). In its more general form~\cite{lawler63qap}, the coefficients can be negative, and so there is no major difference between minimizing and maximizing, and we see that our optimization problem~\eqref{eq:opt} is a special case of this.

\section{Parameters used for SiGMa} \label{ap:params}
We use $\alpha=1/3$ as the graph score tradeoff\footnote{This value of $\alpha$ has the nice theoretical justification that it gives twice much more weight to the linear term than the quadratic term, a standard weighting scheme given that the derivative of the quadratic yields the extra factor of two to compensate.}
in~\eqref{eq:score} and $\beta=0.25$ as the property score tradeoff in~\eqref{eq:static}. We set the string score \texttt{smoothing} term in~\eqref{eq:string} as the sum of the maximum possible word weights in each $\KB$ ($\log |\mathcal{E}_o|$). We use 0.25 as the score threshold for the stopping criterion (step 6 in the algorithm), and stop considering suggestions from the inverted index on strings when their score is below 0.75. We use as initial matching the unambiguous exact string comparison test as described in Section~\ref{sec:algorithm}. We use uniform weights $w_{ik} = 1$ for the matched neighbors contribution in the graph score~\eqref{eq:graphscore}.  We use a \texttt{Sim} measure on property values as used in~\eqref{eq:property} which depends on the type of property literals: for dates and numbers, we simply use $0$-$1$ similarity (1 when they are equal) with some processing --- e.g. for dates, we only consider the year; for secondary strings (i.e. strings for other properties than the main string representation of an entity), we use a weighted Jaccard measure on words as defined in~\eqref{eq:string} but with the IDF weights derived from the strings appearing in this property only.

\section{Additional parameter \\experiments} \label{ap:param_exp}
We provide here the additional parameter experiments which were skipped from the main text for brevity.

\subsection{Exp. 5: Weighting schemes, smoothing and tradeoffs}
\begin{figure}[!t]
	\begin{center}
    \includegraphics[width=\columnwidth]{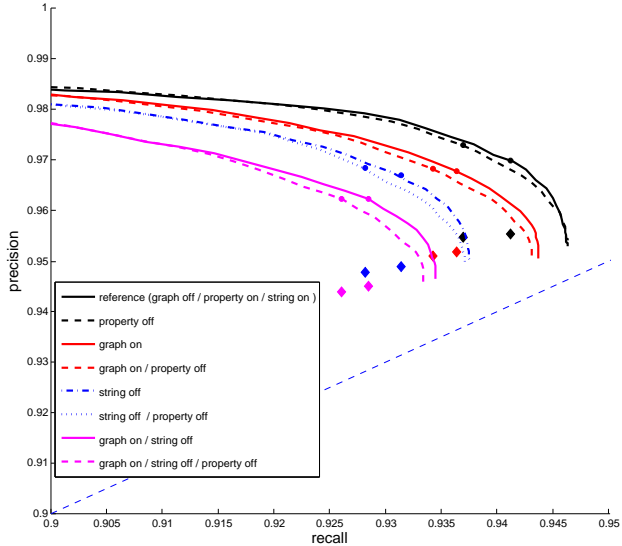}
    \end{center}
	\caption{\textbf{Exp. 5: Precision/Recall curves for \ts{SiGMa} on \ts{YAGO}-\ts{IMDb} with different weighting configurations.} \textnormal{The filled circles indicate the maximum F-measure position on each curve, with the corresponding diamond giving the F-measure value at this recall point. Each curve is one of the 8 possibilities of having the weight `off' (set to unity) or `on', for the graph / property / string part of the score function. The legend indicates the difference between the reference setup (graph off / property on / string on) and the given curve.} \label{fig:test_weights} }
\end{figure}

In this experiment,  we explored the effect of the weighting scheme for the three different score components (string, property and graph) by trying two options per component, with precision / recall curves given in Figure~\ref{fig:test_weights}. For string and property components, we compared uniform weights vs. IDF weights. For the graph component, we compare uniform weights (which surprisingly got the best result) with the inverse number of neighbors weight proposed in~\eqref{eq:w_ik}. Overall, the effect for these variations was much smaller than the one for the score component experiment, with the biggest decrease of less than 1\% F-measure obtained by using uniform string weights instead of the IDF-scores. We also varied the 3 smoothing parameters (one for each score component) as well as the 2 tradeoff parameters linearly around their chosen values: the performance does not change much for changes of the order of 0.1-0.2 for the tradeoff, and 1.5 for the smoothing parameters (stay with 1\% range of F-measure).

\subsection{Exp. 6: Stopping threshold choice}
\begin{figure}[!t]
	\begin{center}
		\includegraphics[width=\columnwidth]{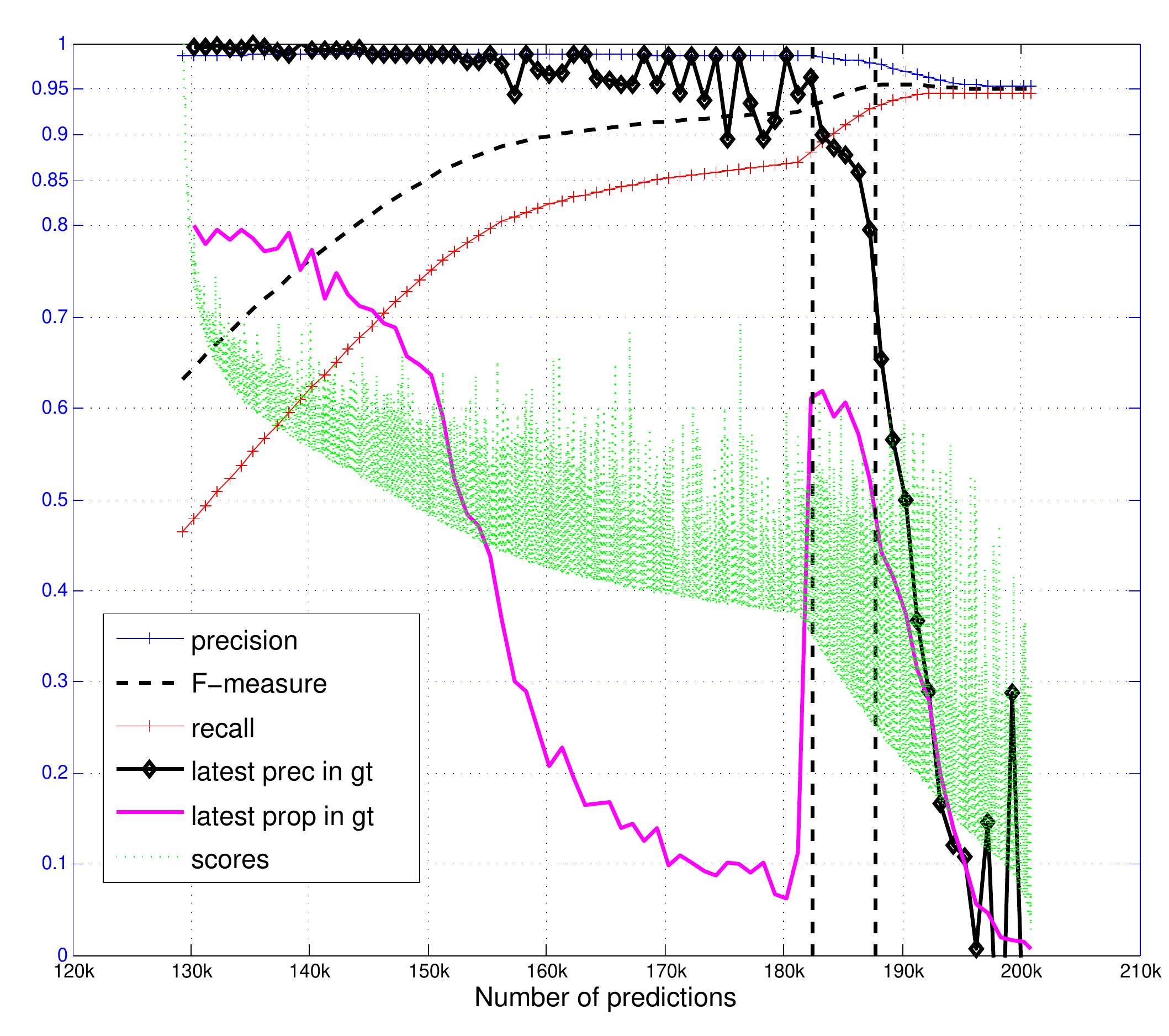}
	\end{center}
	\caption{\textbf{Exp. 6: Precision/recall and score evolution for \ts{SiGMa} on the \ts{YAGO}-\ts{IMDb} dataset as a function of iterations (predictions).} \textnormal{The magenta line indicates the proportion out of the last 1k predictions for which we had ground truth information; the black line with diamonds indicate the precision for these 1k predictions. The score of the matching pair chosen at each iteration is shown in green; notice how the precision starts to drop when the score goes below 0.35 (first vertical black dashed line) and the F-measure starts to level when the score goes below 0.25 (second vertical dashed line). We note that the periodic increase of the score is explained by the fact that if compatible neighbors are matched, the graph score part~\eqref{eq:graphscore} of their neighbors can increase sufficiently to exceed the previous maximum score in the priority queue.
}} \label{fig:detailed_run}
\end{figure}

In this experiment, we studied whether the score information correlated with changes in the precision / recall information, in order to determine a possible stopping threshold. We overlay in Figure~\ref{fig:detailed_run} the precision / recall at each iteration of the algorithm (blue / red) with the score (in green) of the matched pair chosen at this iteration (as given by~\eqref{eq:score}). The vertical black dashed lines correspond to the iteration at which the score threshold of 0.35 and 0.25 are reached, respectively, which correlated with a drop of precision for the current predictions (black line with diamonds) and a leveling of the F-measure (curved dashed black line), respectively. We note that this correlation was also observed on all the other datasets, indicating that this threshold is robust to dataset variations.

\newpage

\mbox{   } %

\end{appendix}

\end{document}